\documentclass{article}

\usepackage[main,final]{neurips_2026}

\usepackage[utf8]{inputenc} 
\usepackage[T1]{fontenc}    
\usepackage[colorlinks]{hyperref}
\usepackage{url}            
\usepackage{booktabs}       
\usepackage{amsfonts}       
\usepackage{nicefrac}       
\usepackage{microtype}      
\usepackage[dvipsnames,table]{xcolor}         

\usepackage{amsthm}
\usepackage{amsmath}
\usepackage{amssymb}
\usepackage{mathtools}
\usepackage{amsthm}
\usepackage{cleveref}
\usepackage{wrapfig}
\usepackage{gradient-text}  
\usepackage{multirow}
\usepackage{enumitem}
\usepackage{pifont}
\usepackage{graphicx}
\usepackage[most]{tcolorbox}
\usepackage{colortbl}
\usepackage{makecell}
\usepackage{array}
\usepackage{siunitx}
\usepackage{xspace}

\newcommand{\colorband}{%
  \raisebox{-0.05cm}{\includegraphics[height=0.25cm,width=0.8cm]{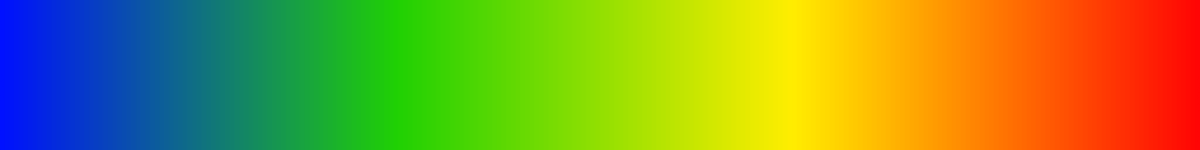}}%
}

\theoremstyle{definition}

\theoremstyle{plain}

\newcommand*{\ie}{\emph{i.e.},\@\xspace}

\definecolor{IllinoisOrange}{HTML}{FF5F05}
\definecolor{IllinoisBlue}{HTML}{13294B}

\definecolor{SoftBeige}{HTML}{FEF5E7}
\definecolor{YaleBlue}{HTML}{2A5487}
\definecolor{CustomRed}{RGB}{233, 93, 34}      
\definecolor{CustomOrange}{RGB}{245, 132, 38}  
\definecolor{CustomPeach}{RGB}{255, 165, 82}   
\definecolor{CustomLightOrange}{RGB}{255, 170, 120} 
\definecolor{CustomLightRedOrange}{RGB}{255, 195, 160} 
\definecolor{CustomYellow}{RGB}{255, 220, 148} 
\definecolor{CustomDarkGray}{RGB}{80, 80, 80}
\definecolor{CustomTeal}{HTML}{008080}   
\definecolor{BrightTeal}{HTML}{00A6A6}
\definecolor{CustomMint}{HTML}{50C99A}
\definecolor{CustomPink}{HTML}{FF828F}
\definecolor{CustomPurple}{HTML}{884AB2}
\definecolor{CustomLightPurple}{HTML}{C497FF}
\definecolor{CustomLightLightPurple}{HTML}{E6DEFF}
\definecolor{CustomPink}{RGB}{255, 138, 128}

\definecolor{SILSABest}{HTML}{EEE7FF}     
\definecolor{SILSASecond}{HTML}{DDF7F2}   
\definecolor{SILSARow}{HTML}{F0F2F7}   
\definecolor{SILSAAccent}{HTML}{0F9B8E}   

\providecommand{\best}[1]{}
\providecommand{\second}[1]{}

\renewcommand{\best}[1]{%
  \cellcolor{SILSABest}\textbf{#1}%
}

\renewcommand{\second}[1]{%
  \cellcolor{SILSASecond}#1%
}
\newcommand{\captionbox}[2]{%
  \begingroup
  \setlength{\fboxsep}{1.2pt}%
  \colorbox{#1}{\strut #2}%
  \endgroup
}

\definecolor{SliceNavy}{HTML}{1B2A59}
\definecolor{SliceTeal}{HTML}{0F9B8E}
\definecolor{SliceMint}{HTML}{DDF7F2}
\definecolor{SliceLavender}{HTML}{EEE7FF}
\definecolor{SlicePurple}{HTML}{6A4C93}
\definecolor{SliceOrange}{HTML}{F28C28}
\definecolor{SliceRose}{HTML}{FCE7EF}
\definecolor{SliceGray}{HTML}{F6F7FB}
\definecolor{SliceGold}{HTML}{F7C948}

\newcommand{\modelnamegradient}
{\textbf{\gradientRGB{SI}{238, 160, 207}{89, 190, 225}\gradientRGB{LSA}{89, 190, 225}{139, 207, 79}}\xspace}

\newcommand{\modelnamenc}{{SILSA}\xspace}
\newcommand{\slicevae}{{\textbf{SliceVAE}}\xspace}

\definecolor{SILSASection}{HTML}{E9EDF5} 

\newtcolorbox{takeawaybox}[1][]{%
  enhanced,
  colback=SliceMint!65,
  colframe=SliceTeal!85!black,
  boxrule=0.55pt,
  arc=2mm,
  left=1.2mm,right=1.2mm,top=0.8mm,bottom=0.8mm,
  fonttitle=\bfseries,
  title={Key Takeaway},
  #1
}

\newtcolorbox{contributionbox}[1][]{%
  enhanced,
  colback=SliceLavender!65,
  colframe=SlicePurple!80!black,
  boxrule=0.55pt,
  arc=2mm,
  left=1.2mm,right=1.2mm,top=0.8mm,bottom=0.8mm,
  #1
}

\title{\modelnamegradient: Sliding-Window Slice Latents for Topology-Preserving High-Resolution 3D Generation}

\author{%
 \textbf{Tianjiao Yu, Xinzhuo Li, Yifan Shen, Ying Shen,}\\\textbf{ Kiet A. Nguyen, Adheesh Sunil Juvekar, Ismini Lourentzou} \\ 
  University of Illinois Urbana-Champaign\\
  \texttt{\{ty41,lourent2\}@illinois.edu} \\[5pt]
  \href{https://plan-lab.github.io/silsa}{\texttt{\textcolor{IllinoisOrange}{plan-lab.github.io/silsa}}}
}

\begin{document}

\maketitle

\begin{figure*}[!h]
\vspace{-0.7cm}
    \centering
    \includegraphics[width=0.99\textwidth]{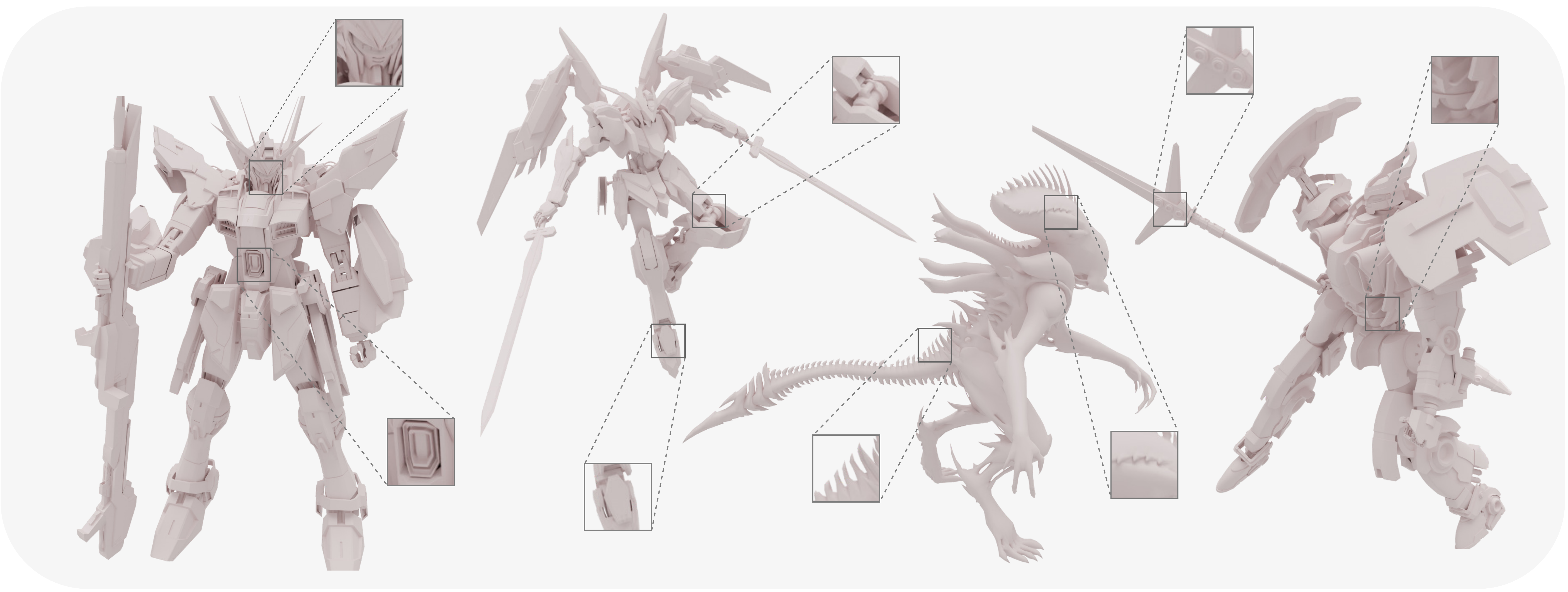}
    \vspace{-0.35cm}
    \caption{\textbf{High-resolution 3D generation through topology-preserving slice latents.} 
    Given a single input image, \modelnamegradient{} generates high-resolution 3D shapes with coherent structure and detailed local geometry, including thin structures, holes, repeated components, and long-range connectivity. By modeling shapes through compact overlapping slice latents and cross-axis volumetric coordination, \modelnamenc{} maintains cross-sectional continuity and produces coherent global structure.}
    \label{fig:teaser}
    \vspace{0.2cm}
\end{figure*}

\begin{abstract}
High-resolution 3D generation increasingly relies on voxel latents and multi-stage pipelines that first predict active structure and then synthesize local geometry. 
While effective, this design fragments continuous surfaces into many local tokens, inflates generation cost, and often weakens topological consistency for thin or highly connected shapes. 
We introduce \modelnamegradient{}, a topology-aware 3D generation framework that represents shapes with compact \emph{sliding-window slice latents}. 
Instead of generating expensive voxel tokens, \modelnamenc{} uses a fixed set of overlapping slices along the three canonical axes, where each token summarizes a local depth window to preserve cross-sectional continuity and support single-stage rectified-flow generation. 
A \textbf{Slice VAE} encodes oriented surface samples into multi-axis slice latents and reconstructs them with a sparse volumetric decoder, while a \textbf{Volumetric Anchor Lattice} coordinates directional slice streams through a shared 3D workspace. 
To encourage structural fidelity, we introduce slice-level topology supervision that matches persistence diagrams and aligns Betti transitions across neighboring slices.
Experiments show that \modelnamenc{} improves structural fidelity while substantially reducing generation cost. 
\modelnamenc{} improves PSNR by $24.4\%$, coverage by $6.07$ absolute points, and Betti error by $9.2\%$ over the strongest baseline, while using $70.0\%$ fewer tokens than the next-most compact baseline and over $98\%$ fewer tokens than sparse or hierarchical tokenizers. Qualitative results further show improved preservation of thin structures, repeated components, and long-range connectivity.\looseness-1
\end{abstract}
\section{Introduction}
High-resolution 3D generation has advanced rapidly from per-instance optimization toward generative models that synthesize complete 3D assets from a single image or text prompt. Early score-distillation and multi-view diffusion pipelines demonstrated that strong 2D generative priors can be lifted into plausible 3D objects~\citep{poole2022dreamfusion,lin2023magic3d,wang2023prolificdreamer,liu2023zero,liu2023syncdreamer,long2024wonder3d,shi2023zero123++}. More recent native 3D generators learn compact latent spaces and train diffusion, autoregressive, or flow models directly over 3D structure~\citep{jun2023shap,zhao2023michelangelo,zhang20233dshape2vecset,zhang2024clay,ren2024xcube,xiang2025structured,he2025sparseflex,yu2026elsa3d,zhao2025hunyuan3d}.
These systems make 3D generation substantially faster and more scalable, but they still struggle to preserve fine 3D structure. Generated shapes often match the overall object appearance while breaking thin parts, openings, and repeated components. These failures alter connectivity, remove openings, and break part relationships.

This is a representation challenge fundamental to high-resolution 3D generation.
Dense voxel grids provide a direct spatial scaffold by representing shape as occupancy or signed-distance values on a regular 3D lattice~\citep{cheng2023sdfusion,wu20153d}, but their memory and computation grow cubically with resolution.
Sparse voxel, octree, and hierarchical tokenizers reduce this cost by modeling only occupied regions, high-detail regions, or progressively refined geometry~\citep{liu2020neural,riegler2017octnet,tatarchenko2017octree,ren2024xcube,xiang2025structured,maruani2025shapeshifter}.
However, their token count can remain data-dependent and may grow for objects with thin supports, many repeated parts, or complex surface topology.
Compact alternatives, including set-based shape latents~\citep{zhang20233dshape2vecset,zhao2023michelangelo,jun2023shap}, triplanes~\citep{chan2022efficient,fridovich2023k,gupta20233dgen,hong2023lrm}, primitive-based representations~\citep{tang2024lgm} keep generation tractable, but they weaken the direct correspondence between a token and the local geometric structure it must preserve.
As a result, a model may achieve low surface or rendering error while still producing a structurally incorrect shape, such as filling a hole, breaking a support, or merging two nearby components.\looseness-1

To address this dilemma, we introduce \modelnamegradient{}, a 3D generation framework that represents high-resolution shapes with \textbf{\emph{sliding-window slice latents}}.
The key insight is that a slice (with a small thickness) captures the topology of an entire planar region in one coherent unit, whereas existing representations either fragment this structure across many local cells or compress it into tokens with weakened spatial correspondence. A sequence of slices further preserves continuity along the slicing axis, directly exposing how connected components, holes, and part boundaries evolve along each canonical direction.
Motivated by these properties, \modelnamenc{} decomposes a shape into three sequences of overlapping slice latents along the $x$, $y$, and $z$ axes.
For $N$ slice positions per axis, the generator operates on only $3N$ latent tokens, regardless of the object's occupancy, surface area, or part complexity.
Because each token summarizes a local depth window instead of an infinitesimal plane, the representation remains compact while still exposing thin parts, nearby surfaces, and small openings to the model.
The three axis-wise sequences provide complementary cross-sectional views of the same object, giving \modelnamenc{} a short, spatially indexed latent sequence for high-resolution 3D generation.\looseness-1

To make this representation effective for generation, \modelnamenc{} combines compact slice latents with cross-axis coordination and topology-aware training.
First, a \slicevae{} encodes oriented surface samples into overlapping slice latents and reconstructs them through a sparse volumetric decoder, preserving local surface geometry.
A \textbf{Volumetric Anchor Lattice} then coordinates the three directional slice streams inside the rectified-flow transformer: each slice token reads from and writes to the anchor plane associated with its axis and depth position, allowing $x$-, $y$-, and $z$-aligned evidence to accumulate in a shared 3D workspace and form a single coherent shape.
Finally, slice-level topology losses supervise the decoded cross-sections by matching persistent-homology structure and aligning Betti transitions across adjacent slices~\citep{edelsbrunner2002topological,zomorodian2004computing,hu2019topology,clough2020topological,stucki2023topologically,stucki2024efficient}.
Together, these components allow \modelnamenc{} to improve both geometric fidelity, such as accurate surfaces and part shapes, and topological fidelity, such as connected components, holes, and consistent connectivity across depth.

Experiments show that \modelnamenc{} improves high-resolution image-to-3D generation while substantially reducing generation cost. 
Across both settings, \modelnamenc{} preserves fine structures that are commonly degraded by compact 3D latents, including thin supports, handles, holes, spokes, railings, and repeated parts.
Empirically, \modelnamenc{} improves both structural fidelity and efficiency. 
On image-conditioned 3D generation, it achieves the best CLIP, FD, PSNR, LPIPS, coverage, and MMD, with a $24.4\%$ relative gain in PSNR and a $6.07$-point absolute gain in coverage over the strongest baselines, while matching the best KD. 
The SliceVAE further reduces Betti error by $9.2\%$ relative to the strongest reconstruction baseline, indicating better preservation of connected components and holes. 
At the same time, \modelnamenc{} uses only $384$ fixed slice tokens, $70.0\%$ fewer than the next-most compact baseline, reducing training memory and inference time. Qualitative results further show that cross-axis coordination through the Volumetric Anchor Lattice reduces inconsistent slice predictions and produces more coherent 3D assets.
In summary, the contributions of our work are:
\begin{itemize}[itemsep=0.5ex, parsep=0pt, topsep=-2.3pt, leftmargin=0.5cm]
    \item We introduce \modelnamegradient{}, a topology-aware image-to-3D generation framework that encodes shapes into a compact set of spatially grounded \textbf{sliding-window slice latents} across three canonical axes.

    \item We design a \textbf{topology-aware SliceVAE} that combines overlapping slice aggregation, sparse volumetric decoding, persistent-homology matching, and Betti-transition supervision to preserve surface geometry, connected components, and hole structures.

    \item We develop a single-stage rectified-flow generator with a \textbf{Volumetric Anchor Lattice}, enabling cross-axis coordination through shared spatial memory and efficient image-conditioned 3D generation from only $384$ slice tokens.
\end{itemize}

\section{Related Work}
High-resolution 3D generation has evolved from lifting 2D diffusion priors through score distillation and differentiable rendering~\cite{poole2022dreamfusion,lin2023magic3d,wang2023prolificdreamer,chen2023fantasia3d} to image-conditioned multi-view reconstruction pipelines~\cite{liu2023syncdreamer,long2024wonder3d,shi2023zero123++,liu2023one,shi2023mvdream} and native 3D generators over learned latent spaces~\cite{cheng2023sdfusion,jun2023shap,zhao2023michelangelo,zhang2024clay,xiang2025structured,xiang2025native,he2025sparseflex,yu2025core3d}. 
Existing 3D representations trade off efficiency and structure: dense voxels provide spatial grounding but scale cubically~\cite{wu20153d,cheng2023sdfusion}, triplanes and set latents improve compactness but weaken local geometric correspondence~\cite{chan2022efficient,fridovich2023k,zhang20233dshape2vecset}, and sparse or hierarchical voxel tokenizers preserve locality but require data-dependent token counts and often multi-stage generation~\cite{ren2024xcube,xiang2025structured,he2025sparseflex}. 
Cross-sectional representations provide a spatially grounded alternative, as planar slices expose components, holes, and connectivity changes that compact global latents can blur, while OReX~\cite{sawdayee2023orex} demonstrates that such slices provide useful geometric cues for reconstruction.
Slice3D~\cite{wang2024slice3d} predicts multi-slice images as intermediate observations for single-view 3D reconstruction. Slice-sequence diffusion has also been explored for paired volumetric medical image and mask synthesis~\cite{han2023medgen3d}.
\modelnamenc{} differs by using multi-axis cross-sections as a learned generative latent with fixed sliding-window slice tokens. 
Moreover, our topology supervision builds on persistent homology and Betti-based losses for preserving connectivity and holes~\cite{edelsbrunner2002topological,zomorodian2004computing,hu2019topology,clough2020topological,stucki2024efficient}, but avoids expensive full-volume topology matching by supervising persistence within slices and Betti transitions across neighboring slices. 
Additional discussion of 3D generation, latent 3D representations, and topology-aware learning is provided in Appendix~\ref{app:related}.\looseness-1

\definecolor{TopoBlue}{HTML}{12AEEA}
\definecolor{TopoRed}{HTML}{C00000}

\section{Method}
State-of-the-art 3D generative models encode shapes into structured latent tokens and generate them with transformer-based diffusion or rectified-flow models~\cite{xiang2025structured, he2025sparseflex}. The token count, however, scales with surface area, making generation expensive and requiring multi-stage pipelines that first predict which voxels are active before generating their structured latents. Beyond efficiency, voxel-level tokenization also fragments continuous surfaces into many local elements, making topological coherence challenging to model.

\begin{figure*}[t]
    \centering
    \includegraphics[width=0.99\textwidth]{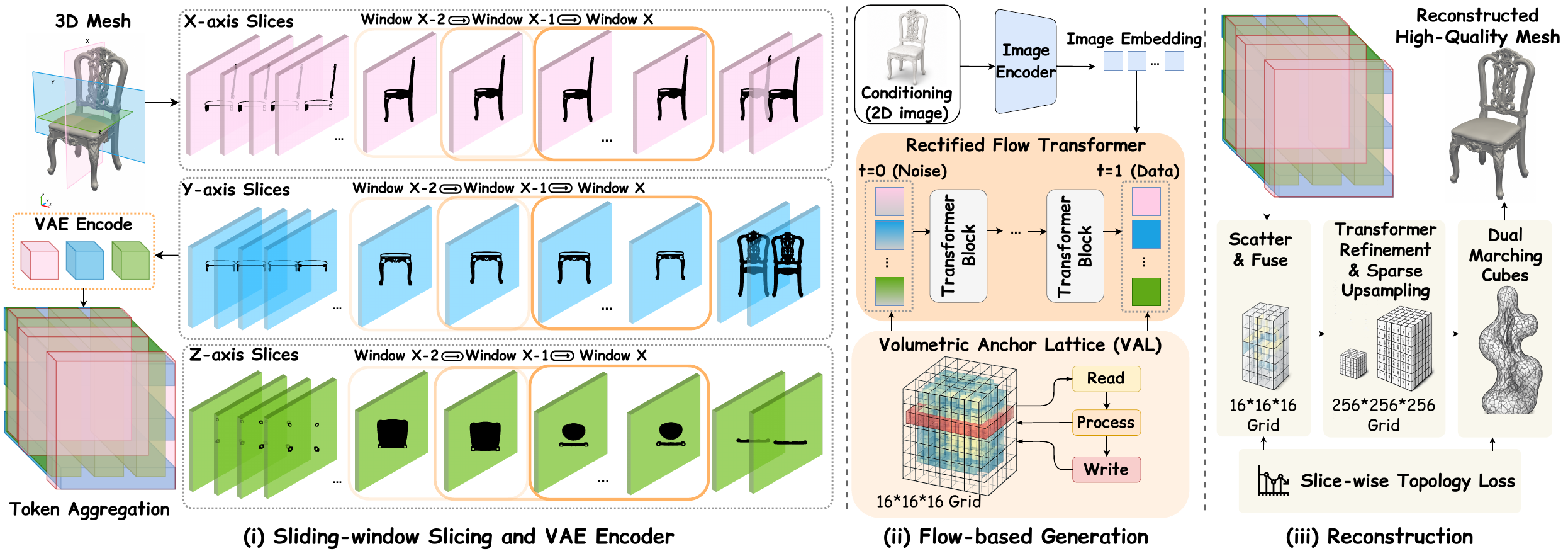}
    \vspace{-0.3cm}
    \caption{\textbf{\modelnamegradient Overview.}
\modelnamenc{} represents each shape with overlapping slice latents along the $x$, $y$, and $z$ axes. A topology-aware SliceVAE encodes surfaces into fixed multi-axis slice latents and decodes them through sparse volumetric upsampling into a high-resolution mesh.
An image-conditioned rectified-flow transformer generates these latents, using a Volumetric Anchor Lattice as shared 3D memory for cross-axis coordination.
Slice-wise persistent-homology and Betti-transition losses supervise connected components, holes, and topological consistency during VAE training.}
    \label{fig:main_method}
    \vspace{-0.3cm}
\end{figure*}

We propose \modelnamenc{} to address these limitations by generating compact, spatially grounded slice latents (\Cref{fig:main_method}).
First, we introduce a \emph{\textbf{SliceVAE}} that maps 3D shapes to compact multi-axis slice latents and decodes them into a high-resolution mesh (\S\ref{sec:vae}). 
The encoder aggregates surface points with overlapping sliding windows along the three canonical axes, yielding a compact set of tokens that preserves local cross-sectional structure. The decoder populates a volumetric feature grid from these tokens and reconstructs geometry through sparse volumetric upsampling. 
To preserve structural correctness, we further introduce a \emph{\textbf{Slice-Wise Topology-Preserving Loss}} that supervises decoded cross-sections (\S\ref{sec:topology}). 
Second, we train a \emph{rectified flow transformer} to generate slice latents from an input image (\S\ref{sec:flow}). Since the slice layout is fixed, generation does not require a separate active-voxel prediction stage. Instead, we introduce a \textbf{\emph{Volumetric Anchor Lattice (VAL)}}, a shared spatial memory that enables slice tokens from different axes to read and write axis-aligned anchor planes during denoising.\looseness-1

\subsection{Topology-Aware Slice VAE}
\label{sec:vae}
\noindent\textbf{Sliding-Window Slice Encoder.}
Inspired by prior work~\cite{he2025sparseflex}, which aggregates point cloud features into sparse voxels via PointNet~\cite{qi2017pointnet}, we adopt the same local pooling paradigm for encoding 3D geometry. However, we replace structured voxels with \(N\) axis-aligned \textit{\textbf{slices}} (planar bins) along each canonical axis. Each slice token summarizes the local geometry within a depth interval, reducing the representation to \(3N\) tokens in total (\(N\) tokens for each of the \(x\), \(y\), and \(z\) axes). The three axis-wise slice sequences provide complementary geometric evidence that the decoder fuses for faithful reconstruction. A single bin, however, may contain very few points. To provide sufficient geometric context, we encode each bin using a \textbf{\textit{sliding window}} of \(w\) surrounding bins.

Formally, given a 3D mesh, we sample a point cloud \(\mathcal{P} = \{\mathbf{p}_\ell\}_{\ell=1}^{N_p}\) with normals \(\{\mathbf{n}_\ell\}_{\ell=1}^{N_p}\) and partition the bounding box into \(N = 128\) bins per axis. For bin \(k\) along axis \(j\), the window gathers all points within \(w/2\) bins on either side:
\begin{equation}
\mathcal{P}_k^j =
\left\{
\mathbf{p} \in \mathcal{P}
\;\middle|\;
k - \tfrac{w}{2}
\leq
\lfloor p^{(j)} \cdot N \rfloor
<
k + \tfrac{w}{2}
\right\},
\quad
k = 0, \ldots, N{-}1,
\end{equation}
with boundary bins clamped to \([0,N{-}1]\). Each point \(\mathbf{p} \in \mathcal{P}_k^j\) is augmented with its depth-relative offset
\(  
\delta(\mathbf{p})\!=\!\lfloor p^{(j)} \cdot N \rfloor - k,
\)
which indicates its displacement from the center bin. A shared MLP processes each augmented point independently, and the window representation is obtained by max-pooling the resulting point features:
\begin{equation}
\mathbf{h}_k^j =
\max_{\mathbf{p} \in \mathcal{P}_k^j}
\mathrm{MLP}_\phi
\left(
\mathbf{p},
\mathbf{n}_{\mathbf{p}},
\delta(\mathbf{p})
\right)
\in \mathbb{R}^d .
\end{equation}
The pooled feature \(\mathbf{h}_k^j\) is mapped to posterior parameters
\((\boldsymbol{\mu}_k^j,\log \boldsymbol{\sigma}_k^j)\), defining a Gaussian slice latent\looseness-1
\begin{equation}
q_\phi(\mathbf{z}_k^j \mid \mathcal{P})
=
\mathcal{N}
\left(
\boldsymbol{\mu}_k^j,
\mathrm{diag}\left((\boldsymbol{\sigma}_k^j)^2\right)
\right).
\end{equation}
During training, the decoder receives latent samples \(\mathbf{z}_k^j \sim q_\phi(\mathbf{z}_k^j \mid \mathcal{P})\). After training, we use the posterior mean \(\boldsymbol{\mu}_k^j\) as the deterministic slice latent for flow training. With \(w\!=\!8\), each bin feature is informed by points spanning 8 consecutive slices. The full latent representation is \(\mathcal{Z} = \{\mathbf{z}_k^j\}\), yielding \(3N\!=\!384\) slice latents. For bins whose entire window is empty, we assign a learned \textit{empty} embedding.\looseness-1

\noindent\textbf{Decoder.}
To reconstruct geometry from the slice latents, we first scatter the three axis-wise latent sequences into a shared coarse 3D feature grid.
Because the slice resolution can be higher than the grid resolution, multiple neighboring slice latents are mapped to the same coarse grid plane. Let
\[
\mathcal{K}_u = \{k \mid \lfloor kD/N \rfloor = u\}
\]
denote the set of slice indices mapped to grid plane \(u\). We aggregate the projected slice latents by normalized summation:
\begin{equation}
\Pi_a(u) \mathrel{+}= 
\frac{1}{|\mathcal{K}_u|}
\sum_{k \in \mathcal{K}_u} W_a \mathbf{z}_k^a,
\label{eq:slice_scatter}
\end{equation}
where \(a \in \{x,y,z\}\) denotes the slice axis, and \(\Pi_a(u)\) denotes the corresponding grid plane, \ie
\(\Pi_x(u)=\mathbf{G}[u,:,:]\), \(\Pi_y(u)=\mathbf{G}[:,u,:]\), and \(\Pi_z(u)=\mathbf{G}[:,:,u]\).

This scatter operation fuses the three axis-wise slice decompositions into a shared volumetric representation. A sparse transformer decoder then refines these features, followed by two self-pruning upsampling stages~\cite{ren2024xcube} that progressively subdivide active cells and prune empty regions, increasing the grid resolution from \(16^3\) to \(256^3\). At the final resolution, a linear head predicts per-cell isosurface parameters, including SDF values, vertex deformations, and interpolation weights. The output mesh is then extracted via differentiable Dual Marching Cubes~\cite{shen2023flexible}.

\noindent\textbf{VAE training.}
The SliceVAE is trained end-to-end with differentiable rendering losses
\( 
\mathcal{L}_{\text{render}} =
\lambda_d \mathcal{L}_d
+
\lambda_n \mathcal{L}_n
+
\lambda_m \mathcal{L}_m,
\)
where \(\mathcal{L}_d\), \(\mathcal{L}_n\), and \(\mathcal{L}_m\) are L1 losses on depth, normal, and silhouette maps, respectively. The latent space is regularized by a KL divergence term:
\begin{equation}
\mathcal{L}_{\text{KL}}
=
\sum_{j} \sum_{k=0}^{N-1}
D_{\text{KL}}
\left(
q_\phi(\mathbf{z}_k^j \mid \mathcal{P})
\,\|\, \mathcal{N}(0,I)
\right).
\end{equation}
Additionally, we propose a slice-wise topology-preserving loss \(\mathcal{L}_{\text{topo}}\) (\S\ref{sec:topology}) that supervises the topological correctness of decoded cross-sections. The full training objective is
\begin{equation}
\mathcal{L}_{\text{VAE}}
=
\mathcal{L}_{\text{render}}
+
\beta_{\text{KL}}\mathcal{L}_{\text{KL}}
+
\lambda_{\text{topo}}\mathcal{L}_{\text{topo}} .
\end{equation}

\subsection{Slice-Wise Topology-Preserving Loss}
\label{sec:topology}
Standard rendering losses capture local surface discrepancies but are often insensitive to structural failures in thin or highly connected shapes, such as bicycle wheels with dense spokes or plants with many branching stems. 
We therefore introduce a slice-wise topology-preserving loss that supervises the topology of decoded cross-sections during VAE training. 
Specifically, we use persistent homology~\cite{zomorodian2004computing,edelsbrunner2002topological} to compare per-slice persistence diagrams and align transitions across neighboring slices.
\looseness-1

\begin{figure}[t]
    
    \centering
    \includegraphics[width=0.99\textwidth]{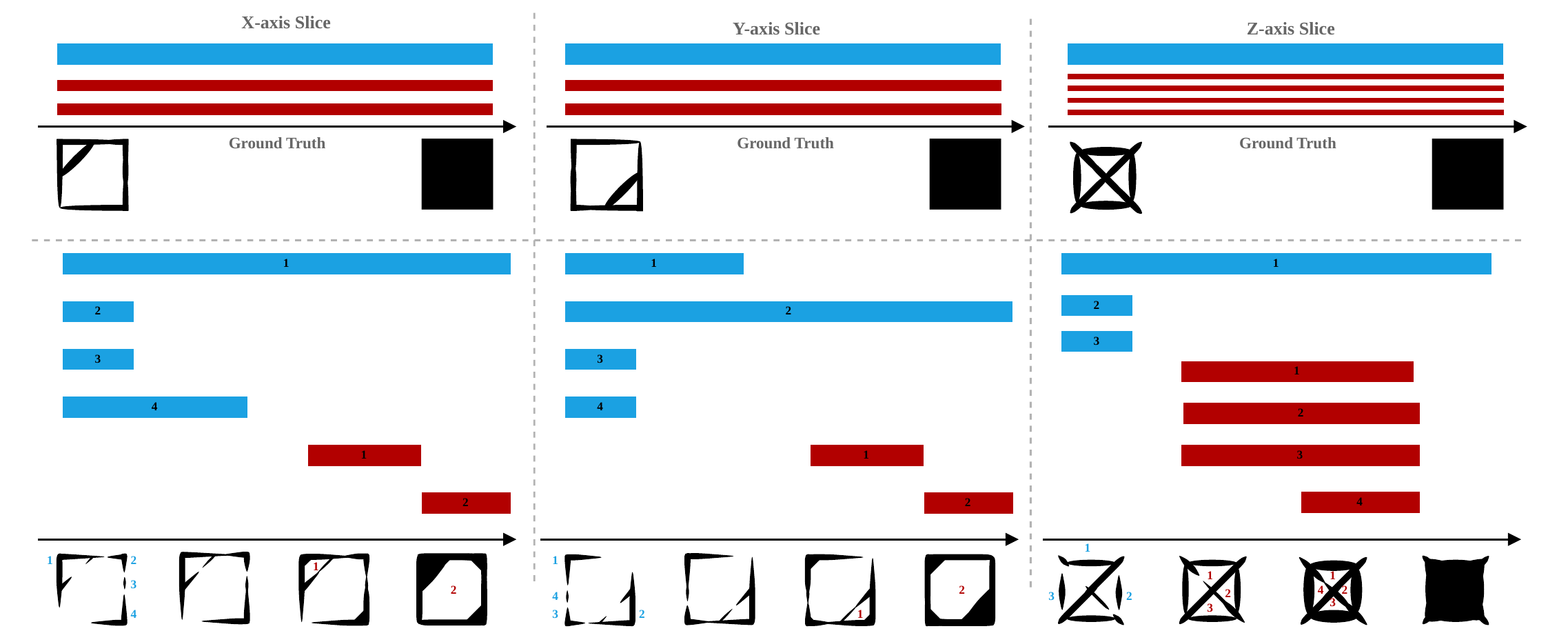}
    \vspace{-0.3cm}
    \caption{\textbf{Topology signals for slice-wise supervision.}
    For each canonical slicing direction, cross-sections form a sequence over depth.
    \protect\textbf{\textcolor{TopoBlue}{Blue}} intervals denote connected components $(\beta_0)$ and 
    \protect\textbf{\textcolor{TopoRed}{red}} intervals denote holes $(\beta_1)$ that persist across ranges of slices.
    Our loss uses these signals in two ways: per-slice persistence matching supervises the topology within each cross-section, while Betti-transition matching supervises where components and holes appear, disappear, merge, or split across neighboring slices.}
    \label{fig:topo}
        \vspace{-0.3cm}
\end{figure}

\noindent\textbf{Illustration of Multi-Axis Slice Topology.} In Figure~\ref{fig:topo}, we visualize how cross-sectional topology evolves as a 3D shape is sliced along the $x$-, $y$-, and $z$-axes. Each column corresponds to one slicing direction, with representative ground-truth cross-sections shown at the top and the induced topological events shown below. 
As the slicing plane moves through the object, each 2D cross-section induces connected components and holes.
\textbf{\textcolor{TopoBlue}{Blue}} intervals track connected components $(\beta_0)$, while 
\textbf{\textcolor{TopoRed}{red}} intervals track holes $(\beta_1)$ over slice depth.
The endpoints of these intervals indicate topological transitions, such as a component appearing, a hole closing, or two regions merging.
Our per-slice persistence loss matches the topology within each decoded cross-section, and our Betti-transition loss aligns the locations of these topological changes across adjacent slices.\looseness-1

After upsampling, the decoder predicts SDF values on a dense corner grid. For efficiency, we compute the topology loss on $N_s$ evenly spaced cross-sections along each canonical axis. For a sampled depth index $k$, we extract a cross-section by indexing the SDF grid and converting to a soft occupancy map $\hat{\mathcal{S}}_k^X(y,z) = \sigma(-\hat{s}[k, y, z] / \kappa)$, with analogous definitions for $\hat{\mathcal{S}}_k^Y$ and $\hat{\mathcal{S}}_k^Z$. 
Here, $\hat{s}$ is the predicted SDF grid and $\kappa>0$ controls the sharpness of the occupancy boundary.
Ground-truth cross-sections are obtained by evaluating signed distances from the target mesh on the same grid and applying the same SDF-to-occupancy conversion.

\noindent\textbf{Topological Signature.}
Each cross-section $\hat{\mathcal{S}}_k^j$ induces a superlevel-set filtration, whose persistence diagram $\mathrm{Dgm}_d(\hat{\mathcal{S}}_k^j)$ records $d$-dimensional topological features as birth--death pairs $(b_p,d_p)$, where $d{=}0$ corresponds to connected components and $d{=}1$ corresponds to holes.
The persistence of a feature is $\mathrm{pers}(p)=b_p-d_p$ under the superlevel convention.
In addition to per-slice persistence diagrams, we compute the Betti number at the occupancy boundary,
\begin{equation}
    B_{k,d}^j=\beta_d\big(\{\hat{\mathcal{S}}_k^j \ge 0.5\}\big),
\end{equation}

and define the transition sequence
\(
\Delta B_{k,d}^j = B_{k+1,d}^j - B_{k,d}^j ,
\)
which captures where cross-sectional topology changes along axis $j$, \ie where connected components or holes appear, disappear, merge, or split as the slicing plane moves through the shape.
By Morse theory, nonzero transitions correspond to intervals containing critical events of the height function along axis $j$~\cite{milnor1963morse}.
We supervise both the per-slice persistence diagrams and the Betti transition sequences against the corresponding ground-truth cross-sections.\looseness-1

\noindent\textbf{Topological Losses.}
We use two complementary losses to supervise the topology of decoded cross-sections.
The \emph{per-slice topology matching} term preserves the topology within each decoded cross-section by matching predicted and ground-truth persistence diagrams:
\begin{equation}
\mathcal{L}_{\text{PH}} =
\sum_{j} \sum_{k=0}^{N_s-1} \sum_{d \in \{0,1\}}
\min_{\gamma \in \Gamma}
\sum_{p \in \mathrm{Dgm}_d(\hat{\mathcal{S}}_k^j)}
\|p-\gamma(p)\|_2^2 ,
\end{equation}
where $j$ indexes the slicing axis, $k$ indexes the sampled cross-section, and $d$ denotes the homology dimension, with $d{=}0$ for connected components and $d{=}1$ for holes. 
The matching set $\Gamma$ includes assignments to ground-truth topological features as well as to the diagonal, so unmatched predicted features are penalized according to their persistence. 
This suppresses spurious short-lived components and holes while preserving persistent structures that define the slice topology.
This matching is asymmetric: unmatched ground-truth features receive no direct persistence-diagram penalty. Missing geometry is additionally constrained by rendering supervision and, when it changes the Betti transition sequence, by the inter-slice transition loss.

The \emph{inter-slice transition matching} term preserves how topology evolves as the slicing plane moves through the shape. 
While per-slice matching encourages each decoded cross-section to have the correct connected components and holes, it does not explicitly couple topology between neighboring slices.
We therefore supervise the Betti transition sequence:
\begin{equation}
\mathcal{L}_{\text{trans}} =
\sum_{j} \sum_{k=0}^{N_s-2} \sum_{d \in \{0,1\}}
\left(
\Delta B_{k,d}^j(\hat{\mathcal{S}})
-
\Delta B_{k,d}^j(\mathcal{S})
\right)^2 .
\end{equation}
Here, \(\Delta B_{k,d}^j = B_{k+1,d}^j - B_{k,d}^j\) records the change in the \(d\)-dimensional Betti number between adjacent slices along axis \(j\). 
Matching these transitions encourages topological events to occur at the correct depths, reducing errors such as holes closing too early, thin supports disconnecting, or nearby parts merging into spurious bridges. 
Since Betti counts are discrete, we compute \(B_{k,d}^j\) from the thresholded occupancy in the forward pass and use a straight-through estimator during backpropagation~\cite{bengio2013ste}.\looseness-1

 The final topology objective combines the two complementary terms:
\begin{equation}
\mathcal{L}_{\mathrm{topo}} =
\lambda_{\mathrm{PH}}\mathcal{L}_{\mathrm{PH}}
+
\lambda_{\mathrm{trans}}\mathcal{L}_{\mathrm{trans}} .
\end{equation}
The persistence term $\mathcal{L}_{\mathrm{PH}}$ preserves the topology of individual cross-sections by matching connected components and holes in persistence-diagram space, while the transition term $\mathcal{L}_{\mathrm{trans}}$ preserves where these structures appear, disappear, split, or merge across neighboring slices.
Together, they encourage the decoded geometry to match both the local topology of each slice and the global evolution of topology along each canonical axis.

\subsection{Rectified Flow Generation with Volumetric Anchors}
\label{sec:flow}
With the SliceVAE trained, we freeze the encoder and decoder and train a rectified flow transformer to generate slice latents from a single input image. The ground-truth latents $\mathcal{Z}^{(0)}$ are obtained by encoding each training shape with the frozen encoder. The transformer learns to map noise to these latents, conditioned on image features.
We use the following flow-matching objective~\cite{lipman2022flow}:
\begin{equation}
\min_\theta \; \mathbb{E}_{t, \boldsymbol{\epsilon}, \mathcal{Z}^{(0)}} 
\left\| 
v_\theta\!\left(\mathcal{Z}^{(t)}, t, \mathbf{c}_{\text{img}}\right) 
- 
(\boldsymbol{\epsilon} - \mathcal{Z}^{(0)}) 
\right\|_2^2,
\end{equation}
where $\mathcal{Z}^{(t)} = (1-t)\mathcal{Z}^{(0)} + t\boldsymbol{\epsilon}$ and $\mathbf{c}_{\text{img}}$ is extracted by a frozen DINOv2 encoder~\cite{oquab2023dinov2}.
 
\noindent\textbf{Volumetric Anchor Lattice (VAL).}
The central challenge in generating multi-axis slice latents is cross-axis consistency: the three axis decompositions must describe one coherent 3D shape, yet each axis-wise sequence is denoised as a separate ordered set of slice tokens. 
We address this challenge with a \emph{Volumetric Anchor Lattice (VAL)}, a persistent 3D feature grid $\mathbf{G} \in \mathbb{R}^{D \times D \times D \times C}$ that serves as shared spatial memory inside the rectified-flow transformer. 
At each transformer block, slice tokens read from and write to axis-aligned planes in this grid according to their slice axis and depth position. 
Tokens from different axes therefore deposit evidence into the same spatial workspace, and later blocks can retrieve this accumulated evidence to coordinate denoising across axes.
This turns cross-axis consistency into a spatially grounded communication mechanism, encouraging the generated $x$-, $y$-, and $z$-aligned slices to decode into one coherent 3D object.

Each slice token has a well-defined spatial footprint in the 
grid: an $x$-slice at depth $k$ maps to the plane 
$\mathbf{G}[k', :, :]$, a $y$-slice to $\mathbf{G}[:, k', :]$, 
a $z$-slice to $\mathbf{G}[:, :, k']$, where $k' = \lfloor kD/N 
\rfloor$. We denote this depth plane as $\mathbf{G}_j(k)$ for 
axis $j$. This correspondence is geometric and requires no 
learning.

\noindent\textbf{Transformer block.}
Each block executes five operations. (1) \emph{Intra-axis Self-attention}: for each axis independently, the $N$ tokens 
attend to each other. (2) \emph{Anchor Read}: each token cross-attends to its depth plane $\mathbf{G}_j(k)$ in the VAL, 
retrieving $D^2$ anchor features that encode what other axes 
have written to the same spatial region. (3) \emph{Anchor 
Write}: each token updates its depth plane via a gated mechanism:\looseness-1
\begin{equation}
\mathbf{G}_j(k) \leftarrow 
(1 - \mathbf{z}_g) \odot \mathbf{G}_j(k) 
+ \mathbf{z}_g \odot f_w(\tilde{\mathbf{h}}_k^j, \mathbf{g}_k^j),
\quad
\mathbf{z}_g = \sigma(W_g[\tilde{\mathbf{h}}_k^j; \mathbf{g}_k^j]),
\end{equation}
where $\tilde{\mathbf{h}}_k^j$ is the token after self-attention, 
$\mathbf{g}_k^j$ is the retrieved anchor feature, 
\(f_w(\tilde{\mathbf{h}}_k^j,\mathbf{g}_k^j)\in\mathbb{R}^{C}\) is broadcast to all cells in the corresponding depth plane, and \(\mathbf{z}_g\in\mathbb{R}^{C}\) is applied channel-wise. (4) \emph{Image Cross-attention} to $\mathbf{c}_{\text{img}}$. (5) \emph{Feed-forward Network}. We stack $L$ blocks; the VAL is initialized to zeros at each velocity evaluation and accumulates cross-axis evidence across transformer blocks.

\begin{table}[t]
  \centering
  \small
  \setlength{\tabcolsep}{8pt}
  \renewcommand{\arraystretch}{1}
\caption{\textbf{Image-to-3D generation.}
We compare \modelnamenc{} with representative image-conditioned and native 3D generation methods. 
\protect\captionbox{SILSABest}{\textbf{Best}} and
\protect\captionbox{SILSASecond}{second best} highlighted.}
  \label{tab:quant_image_to_3d}
  \vspace{-0.25cm}
  \resizebox{\linewidth}{!}{%
  \begin{tabular}{lccccccc}
    \toprule
    \textbf{Model}
    & \textbf{CLIP} $\uparrow$
    & \textbf{FD} $\downarrow$
    & \textbf{KD} $\downarrow$
    & \textbf{PSNR} $\uparrow$
    & \textbf{LPIPS} $\downarrow$
    & \textbf{COV(\%)} $\uparrow$
    & \textbf{MMD(\textperthousand)} $\downarrow$ \\
    \midrule
    Shap-E         & 80.16 & 34.64 & 0.87 & 16.84 & 0.21 & 61.41 & 19.19 \\
    LN3Diff        & 82.79 & 26.98 & 0.76 & 18.73 & 0.19 & 55.21 & 19.84 \\
    Direct3D       & 74.12 & 24.97 & 0.33 & 22.36 & 0.17 & 58.72 & 18.46 \\
    3DTopia-XL     & 76.46 & 24.21 & 0.29 & 22.06 & 0.18 & 58.93 & 17.62 \\
    InstantMesh    & 84.41 & 20.13 & 0.29 & 25.72 & 0.11 & 66.84 & 16.72 \\
    Gau.Any.       & 80.91 & 22.46 & 0.44 & 23.84 & 0.15 & 60.01 & 15.47 \\
    XCube          & 84.91 & 10.32 & \second{0.09} & 23.99 & 0.13 & \second{73.01} & 14.92 \\
    Dora           & 80.35 & 22.84 & 0.23 & 24.68 & 0.13 & 67.42 & 15.63 \\
    SAR3D          & 84.67 & 22.12 & 0.18 & \second{26.31} & \second{0.10} & 70.30 & 15.12 \\
    Trellis        & \second{85.03} & \second{10.31} & \best{0.08} & 24.01 & 0.14 & 72.10 & \second{14.36} \\
    \midrule
    \rowcolor{SILSARow}
    \textbf{\modelnamegradient}
    & \best{87.94}
    & \best{10.16}
    & \best{0.08}
    & \best{32.74}
    & \best{0.05}
    & \best{79.08}
    & \best{14.02} \\
    \bottomrule
  \end{tabular}%
  }
  \vspace{-0.2cm}
\end{table}

\section{Experiments}

\noindent \textbf{Experiment Setup.}
We use Trellis-500K~\cite{xiang2025structured} for training. For evaluation, we use 200 randomly sampled Toys4K assets~\cite{stojanov2021using} and 50 in-the-wild images, with no overlap with the training set. 
We compare the reconstruction quality of Slice VAE with Dora~\cite{chen2025dora}, XCube~\cite{ren2024xcube}, Trellis~\cite{xiang2025structured}, and SparseFlex~\cite{he2025sparseflex}. We use Chamfer Distance (CD) and F-score with thresholds of 0.01 and 0.005, and Betti Error~\cite{stucki2024efficient, hu2019topology} 
to assess geometric fidelity and topological correctness, respectively.
For image-to-3D generation, we compare with representative open-source methods, including Shap-E~\cite{jun2023shap}, LN3Diff~\cite{lan2024ln3diff}, Direct3D~\cite{wu2024direct3d}, 3DTopia-XL~\cite{chen20253dtopia}, InstantMesh~\cite{xu2024instantmesh}, GaussianAnything~\cite{yushi2025gaussiananything}, XCube~\cite{ren2024xcube}, Dora~\cite{chen2025dora}, SAR3D~\cite{chen2025sar3d}, and Trellis~\cite{xiang2025structured}. We use CLIP similarity~\cite{radford2021learning} to evaluate input-output alignment. Overall generative quality is measured with FD~\cite{heusel2017gans} and KD~\cite{binkowski2018demystifying}, while PSNR and LPIPS capture reconstruction-level visual fidelity. We further report COV and MMD~\cite{achlioptas2018learning} to assess distributional fidelity. Full implementation details are provided in Appendix \ref{app:implementation_detail}.

\noindent\textbf{Image-to-3D.}
Table~\ref{tab:quant_image_to_3d} shows that \modelnamenc{} achieves the best CLIP, FD, PSNR, LPIPS, coverage, and MMD among the compared methods, while tying Trellis for the best KD. CLIP increases from 85.03 for Trellis to 87.94, indicating stronger input-image alignment. PSNR reaches 32.74\,dB versus 26.31\,dB for SAR3D, an improvement of 6.43\,dB, while LPIPS decreases from 0.10 to 0.05. These complementary improvements indicate better pixel-level and perceptual fidelity. FD decreases from 10.31 for Trellis to 10.16, while KD remains tied at 0.08. Coverage increases from 73.01\% for XCube to 79.08\%, a gain of 6.07 percentage points, and MMD decreases from 14.36 for Trellis to 14.02. Higher coverage together with lower MMD indicates broader representation of the reference distribution alongside closer shape matching. These results show that the compact fixed-length slice representation supports improvements across both image fidelity and distributional coverage.

\noindent\textbf{VAE Reconstruction Evaluation.}
Table~\ref{tab:quant_vae_recon} evaluates the reconstruction quality of the proposed SliceVAE using geometric, volumetric, and topology-aware metrics. We report Chamfer Distance (CD), F-Score at thresholds $\tau{=}0.01$ and $\tau{=}0.005$, IoU, and Betti-Err, where Betti-Err is computed as the average test-set mismatch in connected components and holes.
\modelnamenc{} consistently outperforms strong 3D tokenizers across all metrics, achieving the lowest CD, highest F-Scores, highest IoU, and lowest Betti-Err. Compared with SparseFlex, the strongest baseline, \modelnamenc{} reduces CD from 0.61 to 0.59, improves F-Score@0.01 from 96.18 to 96.79, improves F-Score@0.005 from 83.62 to 84.03, and increases IoU from 92.54 to 93.01. More importantly, Betti-Err decreases from 1.743 to 1.582, a 9.2\% relative reduction, showing that SliceVAE better preserves connected components and holes rather than only improving surface-level reconstruction fidelity.
\begin{table}[t]
  \setlength{\tabcolsep}{12pt}
  \centering
  \caption{\textbf{VAE reconstruction quality.}
We compare \modelnamenc{} against representative 3D reconstruction tokenizers using geometric, volumetric, and topology-aware metrics.
Betti-Err measures the average mismatch in connected components and holes.
\protect\captionbox{SILSABest}{\textbf{Best}} and
\protect\captionbox{SILSASecond}{second best} results highlighted.
}
  \label{tab:quant_vae_recon}
  \vspace{-0.3cm}
  \resizebox{0.99\textwidth}{!}{%
  \begin{tabular}{l  ccccc}
    \toprule
    \textbf{Model}
    & \textbf{CD} $\downarrow$ 
    & \textbf{F-Score@0.01} $\uparrow$ 
    & \textbf{F-Score@0.005} $\uparrow$ 
    & \textbf{IoU} $\uparrow$ 
    & \textbf{Betti-Err} $\downarrow$ \\
    \midrule
    Dora            & 9.76 & 67.92 & 38.71 & 64.85 & 4.916 \\
    XCube           & 3.21 & 81.74 & 52.06 & 78.13 & 4.382 \\
    Trellis (SLAT)  & 1.12 & 92.83 & 71.45 & 86.29 & 2.871 \\
    SparseFlex      & \second{0.61} & \second{96.18} & \second{83.62} & \second{92.54} & \second{1.743} \\
    \midrule
   \rowcolor{SILSARow}
    \textbf{\modelnamegradient}
    & \best{\textbf{0.59}} & \best{\textbf{96.79}} & \best{\textbf{84.03}} & \best{\textbf{93.01}} & \best{\textbf{1.582}} \\
    \bottomrule
  \end{tabular}%
  }
  \vspace{-0.1cm}
\end{table}

\begin{table}[t]
  \centering
  \small
  \setlength{\tabcolsep}{7pt}
  \renewcommand{\arraystretch}{1.0}
  \caption{\textbf{Efficiency comparison.}
  Token counts for variable-length methods are reported as mean $\pm$ std over the test set.
  VAE memory is measured with batch size 4 on a single A100.
  Inference time is reported end-to-end per shape.
  \protect\captionbox{SILSABest}{\textbf{Best}} and
  \protect\captionbox{SILSASecond}{second best} highlighted.}
  \label{tab:quant_efficiency}
  \vspace{-0.3cm}
  \resizebox{0.99\textwidth}{!}{%
  \begin{tabular}{lccccccc}
    \toprule
    \textbf{Model}
    & \textbf{\#Tokens} $\downarrow$
    & \textbf{Stages} $\downarrow$
    & \textbf{VAE Params} $\downarrow$
    & \textbf{Train Mem.} $\downarrow$
    & \textbf{Inference} $\downarrow$
    & \textbf{VAE} $\downarrow$ \\
    & 
    & 
    & \textbf{(M)}
    & \textbf{(GB)}
    & \textbf{(s/shape)}
    & \textbf{CD}~~\\
    \midrule
    Dora
    & \second{1{,}280}
    & \best{1}
    & {124}
    & \best{14.6}
    & \second{0.82}
    & 9.76 \\

    XCube
    & 64{,}821
    & 2
    & \best{87}
    & 68.4
    & 2.74
    & 3.21 \\

    Trellis
    & 19{,}847 $\pm$ 4{,}312
    & 2
    & 347
    & 42.7
    & 1.93
    & 1.12 \\

    SparseFlex
    & 87{,}453 $\pm$ 18{,}264
    & 2
    & 213
    & 55.4
    & 1.15
    & \second{0.61} \\
    \midrule
    \rowcolor{SILSARow}
    \textbf{\modelnamegradient}
    & \best{384} $(N{=}128)$
    & \best{1}
    & \second{96}
    & \second{17.4}
    & \best{0.34}
    & \best{0.59} \\
    \bottomrule
  \end{tabular}%
  }
  \vspace{-0.3cm}
\end{table}

\noindent \textbf{Efficiency analysis.}
Table~\ref{tab:quant_efficiency} highlights the efficiency advantage of generating fixed slice latents instead of variable active voxel tokens. \modelnamenc{} uses only 384 tokens, which is over 98\% fewer than Trellis, XCube, and SparseFlex. Despite this large reduction in sequence length, \modelnamenc{} achieves the best reconstruction quality. The compact representation also translates directly into lower computational cost in terms of training memory and inference time. Additionally, \modelnamenc{} uses a one-stage generator, making high-resolution generation faster and more predictable.\looseness-1

\begin{wrapfigure}{r}{0.73\textwidth}
    \centering
    \includegraphics[width=\linewidth]{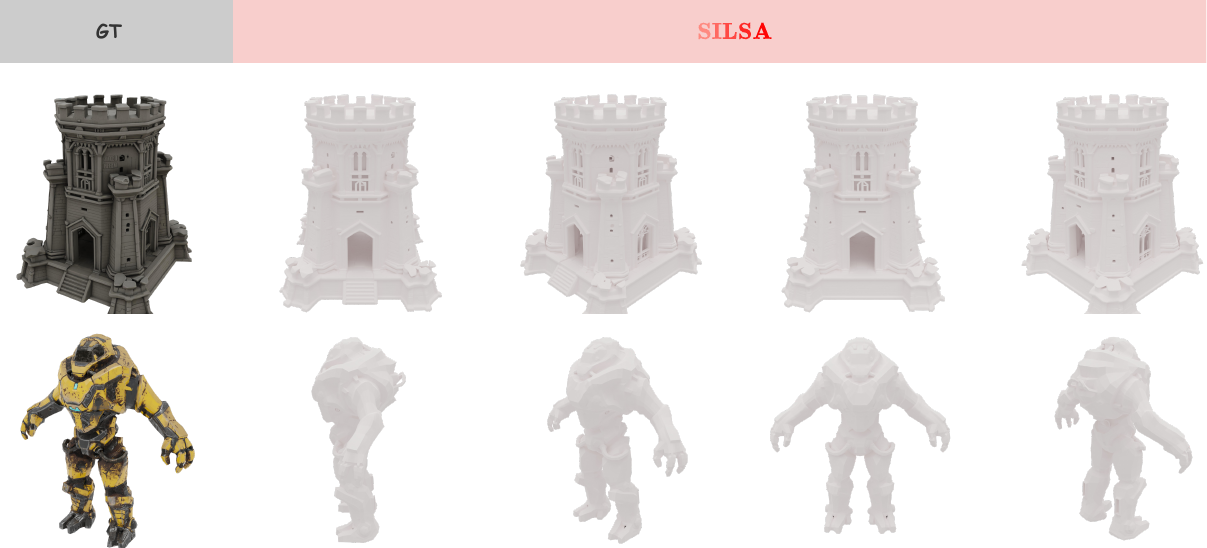}
    \vspace{-0.7cm}
    \caption{\textbf{Image-to-3D generation in the wild.}}
    \label{fig:image_to_3D}
    \vspace{-0.2cm}
\end{wrapfigure}
\begin{figure}[t!]
    \centering
    \includegraphics[width=0.99\linewidth]{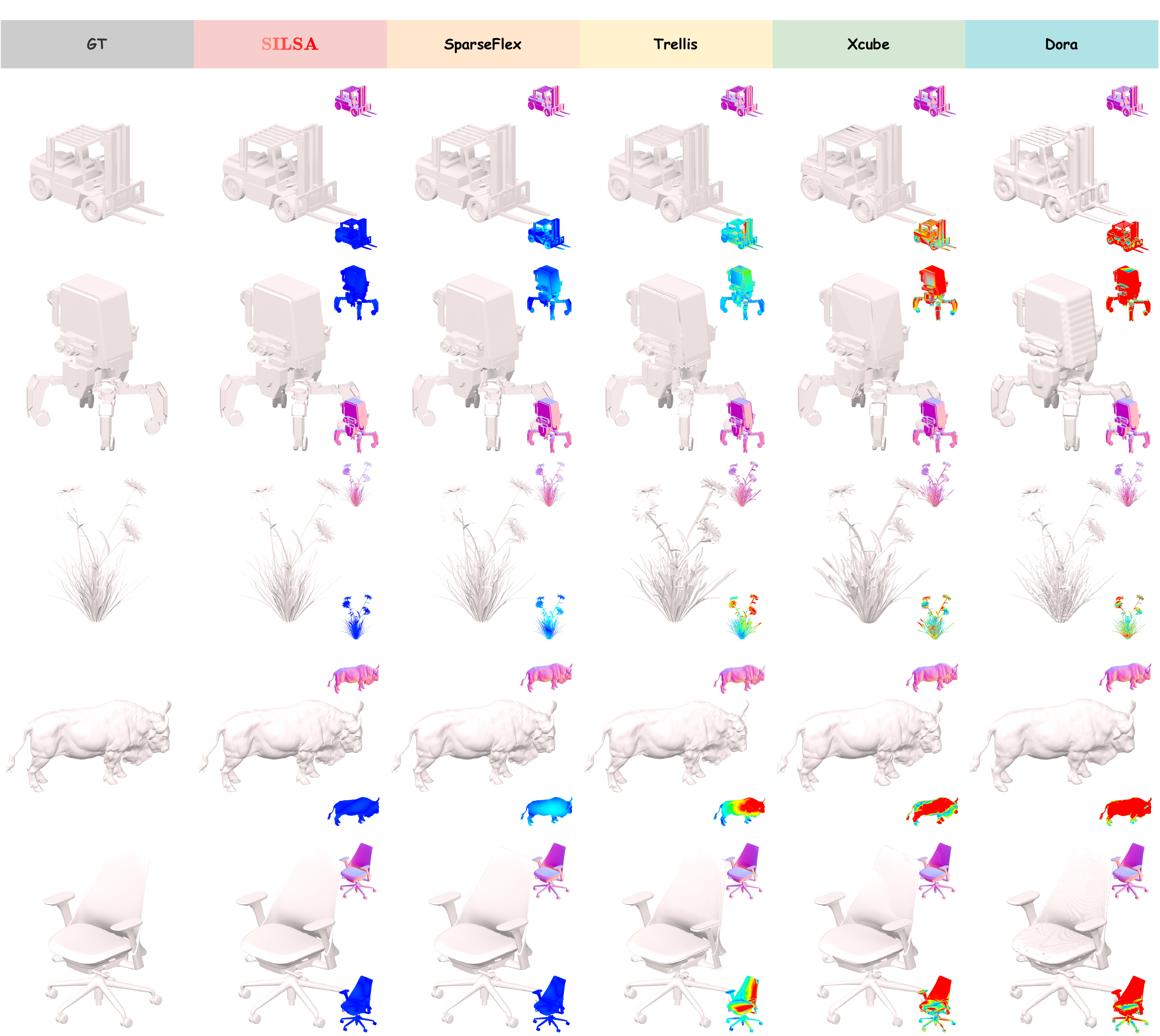}
    \vspace{-0.2cm}
        \caption{\textbf{VAE reconstruction quality.}
    We compare reconstructed meshes from \modelnamenc{} and representative 3D models. Normal maps are shown in the top-right inset, and surface-error maps are shown in the bottom-right inset. Surface error is visualized as \colorband \  from low (blue) to high (red).}
    \label{fig:qual_vae_reconstruction}
\end{figure}

\noindent\textbf{Qualitative Results.}
Qualitatively, \modelnamenc{} better preserves the structural details that are most easily lost in compact 3D latent spaces. For image-to-3D generation, Figure~\ref{fig:image_to_3D} shows that \modelnamenc{} recovers plausible and view-consistent 3D structure from a single image across diverse \emph{in-the-wild} examples. These results indicate that multi-axis sliding-window slice latents preserve local cross-sectional structure while maintaining global coherence.
For VAE reconstruction, Figure~\ref{fig:qual_vae_reconstruction} shows that \modelnamenc{} stays closer to the ground-truth geometry on objects with thin supports, articulated parts, dense branches, holes, and repeated structures, while competing tokenizers often smooth fine details, merge nearby components, or distort fragile parts. Additional examples can be found in Appendix \ref{app:qualitative}.

\newpage
\begin{wraptable}{r}{0.43\textwidth}
  \vspace{-0.2cm}
  \centering
  \scriptsize
  \setlength{\tabcolsep}{5pt}

  \caption{\textbf{Key ablations.}
  We ablate topology supervision (VAE reconstruction) and VAL (image-conditioned generation with the frozen SliceVAE decoder). Highlighted
\protect\captionbox{SILSABest}{\textbf{best}} and
\protect\captionbox{SILSASecond}{second best} results.}
  \label{tab:abl_key}
  \resizebox{\linewidth}{!}{%
  \begin{tabular}{lccc}
    \toprule
    \textbf{Variant}
    & \textbf{CD} $\downarrow$
    & \textbf{IoU} $\uparrow$
    & \textbf{Betti} $\downarrow$ \\
    \midrule

    \rowcolor{SILSASection}
    \multicolumn{4}{l}{\textbf{\textit{Slice-wise topology loss}}} \\
    No $\mathcal{L}_{\mathrm{topo}}$
    & 0.78 & 81.76 & 4.43 \\
    Only $\mathcal{L}_{\mathrm{PH}}$
    & 0.72 & 84.41 & 2.91 \\
    Only $\mathcal{L}_{\mathrm{trans}}$
    & 0.69 & 87.16 & 2.87 \\
    \rowcolor{SILSARow}
    Full $\mathcal{L}_{\mathrm{topo}}$
    & \best{0.59} & \best{93.01} & \best{1.58} \\
    \midrule

    \rowcolor{SILSASection}
    \multicolumn{4}{l}{\textbf{\textit{Cross-axis communication}}} \\
    No communication & 6.21 & 48.92 & 23.41 \\
    Direct cross-attn.
    & \second{0.74} & \best{89.13} & \second{1.89} \\
    \rowcolor{SILSARow}
    VAL
    & \best{0.72} & \second{89.11} & \best{1.81} \\
    \midrule

    \rowcolor{SILSASection}
    \multicolumn{4}{l}{\textbf{\textit{VAL resolution}}} \\
    $D{=}8$
    & 0.79 & 87.02 & 2.06 \\
    \rowcolor{SILSARow}
    $D{=}16$ (default)
    & \best{0.72} & \second{89.11} & \best{1.81} \\
    $D{=}32$
    & \second{0.73} & \best{89.21} & \second{1.83} \\
    \bottomrule
  \end{tabular}%
  }
  \vspace{-0.3cm}
\end{wraptable}

\noindent\textbf{Ablations.}
Table~\ref{tab:abl_key} validates the importance of spatially grounded cross-axis communication. When the three slice streams are generated independently, performance degrades substantially, with CD increasing to 6.21, IoU dropping to 48.92, and Betti-Err rising to 23.41. Direct cross-attention between axes improves consistency, reducing CD to 0.74 and Betti-Err to 1.89 while increasing IoU to 89.13. However, VAL achieves the best geometric and topological fidelity, obtaining the lowest CD of 0.72 and lowest Betti-Err of 1.81, showing that a shared volumetric workspace provides more reliable coordination than unconstrained token-to-token attention. The VAL resolution ablation further shows that $D{=}16$ gives the best trade-off: compared with $D{=}8$, it reduces CD from 0.79 to 0.72, improves IoU from 87.02 to 89.11, and lowers Betti-Err from 2.06 to 1.81. Increasing the resolution to $D{=}32$ slightly improves IoU to 89.21, but worsens CD to 0.73 and Betti-Err to 1.83, suggesting that finer anchors add communication cost without improving structural correctness. The full topology loss gives the best reconstruction trade-off, reducing Betti-Err from 4.43 to 1.58 relative to removing $\mathcal{L}_{\mathrm{topo}}$ while also improving CD from 0.78 to 0.59. Additional ablations in Appendix \ref{app:ablations}.

\begin{wraptable}{r}{0.6\textwidth}
  \vspace{-0.6cm}
  \setlength{\tabcolsep}{5pt}
  \centering
  \small
  \caption{\textbf{Ablation on slice representation.} 
  We ablate the number of slices per axis $N$ and the sliding-window width $w$.
  Default settings are $N{=}128$, $w{=}8$, and three canonical axes, yielding $3N{=}384$ slice tokens. Default settings are shaded.}
  \label{tab:abl_slice}
  \resizebox{\linewidth}{!}{%
  \begin{tabular}{l c  cccc}
    \toprule
    \textbf{Variant} 
    & \textbf{\#Tokens} 
    & \textbf{CD} $\downarrow$ 
    & \textbf{F@0.01} $\uparrow$ 
    & \textbf{IoU} $\uparrow$ 
    & \textbf{Betti-Err} $\downarrow$ \\
    \midrule

    \rowcolor{SILSASection}
    \multicolumn{6}{@{}l}{\textbf{\textit{Number of slices $N$}}} \\
    $N{=}64$    
    & 192  & 0.82 & 94.36 & 88.24 & 2.86 \\
    $N{=}96$    
    & 288  & 0.67 & 95.81 & 91.27 & 2.04 \\
    \rowcolor{SILSARow}
    $N{=}128$ (default) 
    & 384  & 0.59 & 96.79 & 93.01 & 1.58 \\
    $N{=}256$   
    & 768  & 0.57 & 97.02 & 93.28 & 1.53 \\
    $N{=}512$   
    & 1536 & 0.55 & 97.24 & 93.51 & 1.63 \\
    $N{=}1024$  
    & 3072 & 0.56 & 97.18 & 93.42 & 1.77 \\

    \midrule
    \rowcolor{SILSASection}
    \multicolumn{6}{@{}l}{\textbf{\textit{Window width $w$ at $N{=}128$}}} \\
    $w{=}1$ (no overlap) 
    & 384 & 0.73 & 94.18 & 86.95 & 3.24 \\
    $w{=}2$              
    & 384 & 0.68 & 95.09 & 89.32 & 2.46 \\
    $w{=}4$              
    & 384 & 0.63 & 96.14 & 91.85 & 1.91 \\
    \rowcolor{SILSARow}
    $w{=}8$ (default) 
    & 384 & 0.59 & 96.79 & 93.01 & 1.58 \\
    $w{=}16$             
    & 384 & 0.62 & 96.26 & 92.04 & 1.83 \\

    \midrule
    \rowcolor{SILSASection}
    \multicolumn{6}{@{}l}{\textbf{\textit{Window width $w$ at $N{=}256$}}} \\
    $w{=}1$ (no overlap) 
    & 768 & 0.64 & 95.67 & 90.18 & 2.35 \\
    $w{=}2$              
    & 768 & 0.61 & 96.34 & 91.47 & 2.03 \\
    $w{=}4$              
    & 768 & 0.58 & 96.86 & 92.92 & 1.67 \\
    \rowcolor{SILSARow}
    $w{=}8$              
    & 768 & 0.57 & 97.02 & 93.28 & 1.53 \\
    $w{=}16$             
    & 768 & 0.59 & 96.74 & 92.80 & 1.61 \\
    \bottomrule
  \end{tabular}%
  }
  \vspace{-0.4cm}
\end{wraptable}

\noindent\textbf{Slice representation.} Table~\ref{tab:abl_slice} studies how the slice resolution and sliding-window width affect VAE reconstruction. Increasing the number of slices improves reconstruction quality from $N{=}64$ to $N{=}256$, since finer slice bins expose more local geometry and reduce topological ambiguity. However, the gains saturate beyond $N{=}128$: $N{=}256$ and $N{=}512$ slightly improve CD and IoU, but require $2\times$ and $4\times$ more tokens, while $N{=}1024$ further increases token count and worsens Betti-Err, suggesting that overly fine slicing fragments cross-sectional evidence. We therefore use $N{=}128$ as the default because it provides the best efficiency-fidelity trade-off with only 384 tokens. The window-width ablation shows that overlap is critical: $w{=}1$ and $w{=}2$ lack sufficient context and produce higher Betti error, while $w{=}8$ gives the strongest overall balance. Increasing the window to $w{=}16$ slightly degrades performance, likely because excessive aggregation smooths local structures and weakens slice-level specificity.

\section{Conclusion}
We introduce \modelnamegradient{}, a topology-aware image-to-3D framework for high-resolution image-to-3D generation that represents shapes with compact sliding-window slice latents along the three canonical axes. A SliceVAE reconstructs high-resolution geometry with slice-wise persistence and Betti-transition supervision, while a Volumetric Anchor Lattice coordinates directional slice streams inside a single-stage rectified-flow generator. Across reconstruction, generation, and efficiency evaluations, \modelnamenc{} improves geometric fidelity, preserves thin and highly connected structures, and substantially reduces token count, memory, and inference cost. These results show that cross-sectional slice latents offer a compact and structurally faithful representation for scalable 3D generation.

\begin{ack}
This research was partially supported by the NSF CAREER Award \#2542328, Google, the Google TPU Research Cloud (TRC) program, and the Lambda Research Grant. The views and conclusions contained herein are those of the authors and should not be interpreted as necessarily representing the official policies, either expressed or implied, of NSF, Google, Lambda or the U.S. Government. The U.S. Government is authorized to reproduce and distribute reprints for governmental purposes notwithstanding any copyright annotation therein. 
\end{ack}

\bibliographystyle{plain}
\bibliography{neurips}

\appendix
\newpage

\section{Related Work}\label{app:related}

\noindent \textbf{3D Generation.}
Early 3D generation methods commonly adapt pretrained 2D diffusion models to optimize each target asset through differentiable rendering or score distillation~\cite{laine2020modular,poole2022dreamfusion,lin2023magic3d,wang2023prolificdreamer,chen2023fantasia3d,liu2023zero}. These methods reduce the need for large-scale 3D supervision, but they are often heavy in optimization and may inherit multi-view inconsistency from image priors~\cite{liu2023syncdreamer,shi2023mvdream}. 
For image-conditioned generation, multi-view diffusion and reconstruction systems improve consistency by predicting view-consistent observations before 3D reconstruction~\cite{wang2026one2scene,liu2023syncdreamer,long2024wonder3d,shi2023zero123++,zhao2024flexidreamer,liu2023zero,weng2023consistent123,liu2023one,wu2024unique3d,chen2024v3d,voleti2024sv3d,ye2024stablenormal,liu2024reconxreconstructscenesparse}. Related work also provides metric-depth priors~\cite{wang2026depthmaster} and multi-view-consistent editing~\cite{liyi2026omni,chen2026fast}. To improve scalability, later methods learn generative priors directly over compact 3D representations, including implicit fields, SDF latents, point or set latents, triplanes, and aligned image-text-shape latent spaces~\cite{ye2025hi3dgen,li2024craftsman,huang2025spar3d,zhao2023michelangelo,wang2023rodin,cheng2023sdfusion,jun2023shap,zhang20233dshape2vecset,chen2023single,yu2026dreampartgen}. Feed-forward reconstruction models use triplanes~\cite{hong2023lrm} or Gaussian features~\cite{tang2024lgm,zhang2026sparsesplat}. Recent systems further scale native 3D diffusion or rectified-flow transformers over learned latent tokens and sparse structured grids for high-quality conditioned asset generation~\cite{zhang2024clay,xiang2025structured,zhao2025hunyuan3d,he2025sparseflex}. Beyond 3D, learned representations and generative models support image generation, restoration, and editing~\cite{tang2026learning,rombach2022high,brooks2023instructpix2pix,pmlr-v306-susladkar26a,rewardflow2026,pmlr-v306-shen26w,chi2026balancing,chi2026enhancing_flow,zhang2023camouflaged,zhou2026clear,wu2026attention,wu2026look}, video generation~\cite{shen2026phantom,susladkar2026pyratok,shah2026graphvid,shen2026egoforge,guo2026memorize,liu2026driveva,liu2026universe}, and even data augmentation~\cite{zhang2026active,salarian2025medequalizer}. For 3D generation, these advances reflect a broader shift toward representations that balance generative scalability, geometric fidelity, and efficient spatial structure~\cite{ren2024xcube,xiang2025structured,he2025sparseflex,xiang2025native,zhang20233dshape2vecset}.
Building on this trend, \modelnamenc represents geometry as canonical slices along multiple axes, producing compact and spatially grounded latent tokens without explicitly predicting active voxels. \looseness-1

\noindent \textbf{Latent Representations for 3D Shapes.}
Designing compact but expressive representations is central to scalable 3D modeling~\cite{park2019deepsdf,mescheder2019occupancy,chan2022efficient,zhang20233dshape2vecset,xiang2025structured,chen2025dora,yu2025core3d,xu2025hgs,wu2024hgs,wu2025vings}. 
Dense voxel grids provide explicit spatial structure but scale cubically with resolution~\cite{wu20153d,choy20163d}, motivating factorized representations such as triplanes and higher-dimensional plane decompositions, which encode 3D structure through axis-aligned feature planes rather than full volumetric grids~\cite{chan2022efficient,fridovich2023k,gupta20233dgen}.
Another line of work represents shapes with continuous implicit fields~\cite{wu2024swift}, including SDFs and occupancy functions, or compresses them into compact generative latents, volumetric codes, and unordered token sets~\cite{park2019deepsdf,mescheder2019occupancy,chen2019learning,cheng2023sdfusion,jun2023shap,zhang20233dshape2vecset,zhao2023michelangelo,zhang2024clay,chen2025dora}. Compact generative latents improve efficiency, but do not necessarily provide an explicit correspondence between individual tokens and local geometry. Gaussian representations can also be compacted through simplification~\cite{zhang2025gaussianspa}. Part-aware representations model object structure using articulated Gaussian components or hybrid Gaussians and superquadrics~\cite{yu2025part,gao2024partgs}.
A complementary direction preserves spatial locality through sparse or hierarchical voxels, reducing memory by modeling only occupied, adjacent, or progressively refined regions~\cite{riegler2017octnet,tatarchenko2017octree,ren2024xcube,xiang2025structured,he2025sparseflex,xiang2025native}.
More broadly, existing representations navigate a fundamental trade-off between compactness, spatial correspondence, and the computational cost of representing detailed 3D structure~\cite{wang2025s,zhang2026rald,wang2025sddiff,wang2026rauf,li2026mitigating,fan2026detail,ma2026cross,ding2026displacement,zhou2024reducing,zhou2025isosnet,zhang2026automated,zhang2026versaq,zhang2026gaussians}. 
Our representation differs by using axis-aligned slices not as external observations, but as a learned generative latent: a compact $3N$-token layout that combines the spatial grounding of plane-based features with the bounded token count of set-based representations. \looseness-1

\noindent \textbf{Broader Context.} Beyond 3D shape modeling, spatially grounded representations also support robotic manipulation~\cite{liu2026palm}, visual--LiDAR odometry~\cite{liu2026streamvlo}, oriented object detection~\cite{yu2025wholly}, and trajectory prediction from incomplete or noisy observations~\cite{zhang2023layout,zhang2024oostraj}. Efficient reuse of pretrained visual representations is also studied through visual prompt tuning~\cite{xiao2025visual,xiao2026layer} and continual or black-box model adaptation~\cite{zhang2025dpcore,zhang2026adapting,zhang2026prime}. Other work examines visual grounding in multimodal reasoning~\cite{xiao2026rethinking,xiao2026staying} and step-wise supervision for LLM reasoning~\cite{xie2026step}.

\noindent \textbf{Topology-Aware Learning.}
Topology-aware learning uses algebraic-topology tools, especially persistent homology~\cite{edelsbrunner2002topological,zomorodian2004computing,gabrielsson2020topology}, to supervise structural properties that are poorly captured by point-wise or pixel-wise losses. In segmentation, persistent-homology and Betti-based objectives encourage predictions to match target connectivity and hole structure, while skeleton-based losses such as clDice provide efficient topology-preserving surrogates for curvilinear objects~\cite{hu2019topology,clough2020topological,shit2021cldice,byrne2022persistent,stucki2023topologically,hu2021topology}. Shape and topology have also been jointly optimized for constrained 3D design~\cite{christiansen2015combined}. Beyond output supervision, topology has also been used to regularize learned manifolds, as in Topological Autoencoders~\cite{moor2020topological}. Recent work extends these ideas to 3D, including efficient Betti matching for volumetric segmentation, topology-constrained neural implicit reconstruction, and topology-aware reconstruction losses~\cite{stucki2024efficient,jignasu2024stitch,zhang2025topology,shen2025topology,hu2024topology}. However, applying full volumetric persistent-homology supervision inside high-resolution generative training remains expensive, especially when topology must be evaluated repeatedly across decoded samples. \modelnamenc makes topology supervision tractable by aligning the representation with axis-aligned cross-sections: persistent diagrams are matched within individual slices, while Betti transitions are matched across neighboring slices to preserve where topological events occur along each canonical axis.

\section{Implementation Details}
\label{app:implementation_detail}
We train on Trellis-500K~\cite{xiang2025structured}, curated from ObjaverseXL~\cite{deitke2023objaverse}, ABO~\cite{Collins2022abo}, 3DFUTURE~\cite{fu20213d}, and HSSD~\cite{khanna2024habitat}. Each mesh is normalized to a unit bounding box and encoded from $N_p{=}200\mathrm{K}$ oriented surface samples. Unless otherwise stated, \modelnamenc uses $N{=}128$ slice bins per axis, sliding-window width $w{=}8$, and latent dimension $C{=}512$, yielding $3N{=}384$ slice tokens per shape. The decoder scatters latents into a $16^3$ grid, applies sparse refinement and self-pruning upsampling to $256^3$, and extracts meshes with differentiable Dual Marching Cubes~\cite{shen2023flexible}. The Slice VAE is trained for 5 epochs with AdamW, learning rate $1{\times}10^{-4}$, batch size 16, and 8 NVIDIA A100 GPUs. We set $\lambda_d{=}1.0$, $\lambda_n{=}0.5$, $\lambda_m{=}1.0$, $\beta_{\mathrm{KL}}{=}10^{-4}$, and $\lambda_{\mathrm{topo}}{=}0.1$. For topology supervision, we use $N_s{=}128$ sampled cross-sections per axis and SDF-to-occupancy temperature $\kappa{=}0.02$.
The rectified-flow transformer is conditioned on frozen DINOv2 image features~\cite{oquab2023dinov2}. It contains 24 layers, hidden dimension 1024, 16 attention heads, and a VAL resolution of $D{=}16$. We train for 300K steps using AdamW with learning rate $2{\times}10^{-4}$, cosine decay, batch size 256, and 8 NVIDIA A100 GPUs. At inference, we use 50 Euler steps and decode the predicted slice latents with the frozen VAE decoder.

\section{Additional Results}\label{app:qualitative}

\begin{table}[t!]
\centering
\caption{\textbf{VAE reconstruction on open-surface shapes.}  \best{Best} and \second{second best} highlighted.}
\label{tab:vae_deepfashion3d}
\vspace{-0.2cm}
\small
\setlength{\tabcolsep}{10pt}
\renewcommand{\arraystretch}{1.05}
  \resizebox{0.99\textwidth}{!}{%
\begin{tabular}{lccccc}
\toprule
\textbf{Model} & \textbf{CD} $\downarrow$ & \textbf{F-Score@0.01} $\uparrow$ & \textbf{F-Score@0.005} $\uparrow$ & \textbf{IoU} $\uparrow$ & \textbf{Betti-Err} $\downarrow$ \\
\midrule
Trellis (SLAT)   & 0.07 & \second{99.71}  & 91.18 & 96.84 & \second{0.01} \\
SparseFlex       & \second{0.05} & \best{100.00} & \second{93.07} & \second{98.42} & \second{0.01} \\
\rowcolor{SILSARow}
SILSA            & \best{0.04} & \best{100.00} & \best{93.21} & \best{98.71} & \best{0.00} \\
\bottomrule
\end{tabular}
}
\vspace{-0.2cm}
\end{table}

\noindent\textbf{Open Surface Evaluation.}
We further evaluate SliceVAE on the open-surface dataset DeepFashion3D~\cite{zhu2020deep}. As shown in Table~\ref{tab:vae_deepfashion3d}, SILSA achieves the best reconstruction quality across geometric metrics, reducing CD from 0.05 to 0.04 over SparseFlex while matching its perfect F-Score@0.01 and improving F-Score@0.005 from 93.07 to 93.21. 
Topology-aware metrics saturate on this dataset because garments are topologically simple, so all methods that recover the rough surface achieve near-zero Betti-Err. The geometric improvements demonstrate that the slice-latent representation generalizes to open-surface shapes. 

\begin{figure}
    \centering
    \includegraphics[width=0.99\linewidth]{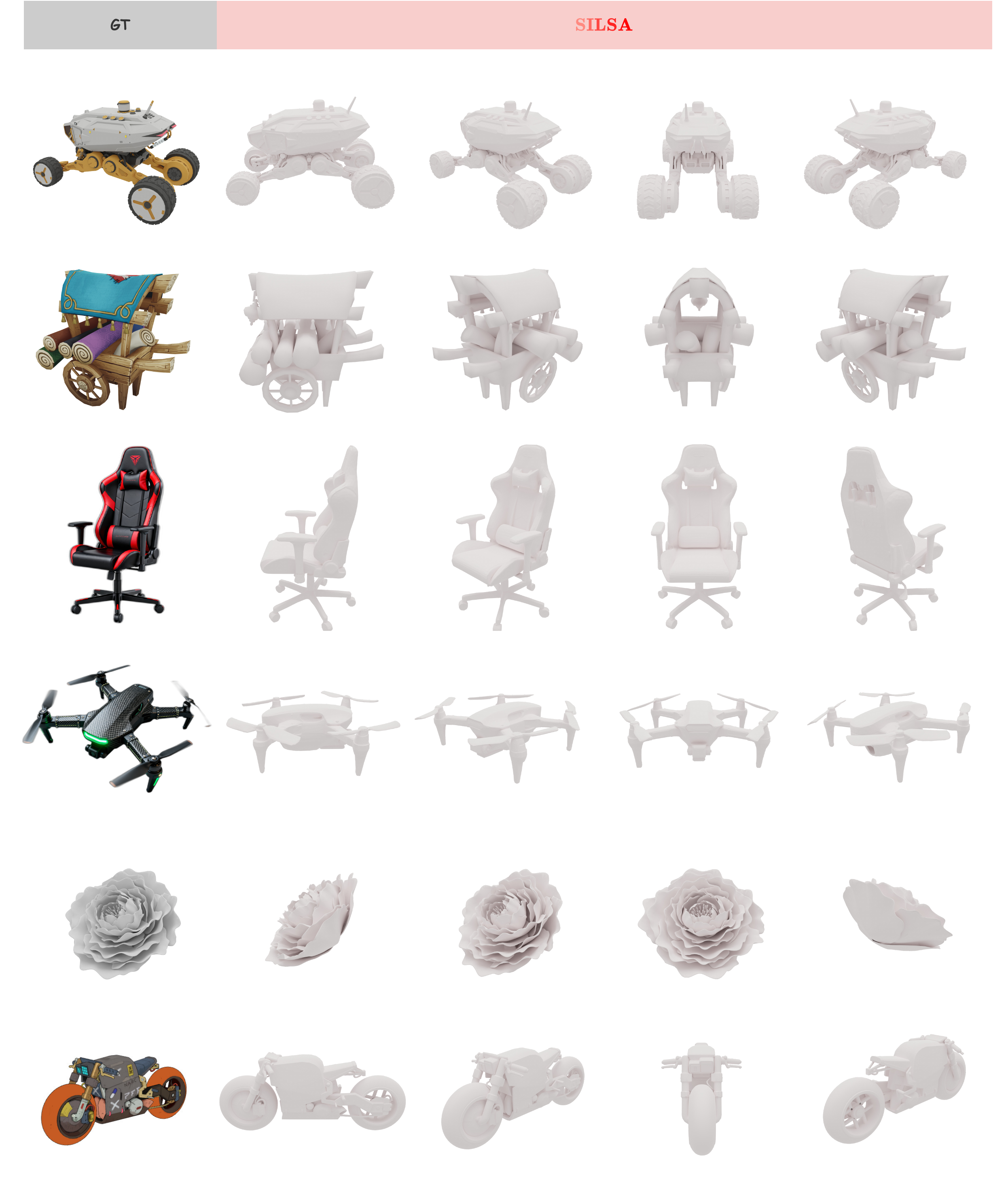}
    \vspace{-0.5cm}
    \caption{\textbf{Additional image-to-3D results.}}
    \label{fig:qual_image_to_3D_supp}
\end{figure}

\noindent\textbf{Image-to-3D in the Wild.}
Figure~\ref{fig:qual_image_to_3D_supp} shows image-conditioned generations on diverse in-the-wild examples. \modelnamenc{} recovers plausible 3D structure from a single view and maintains consistency across rendered viewpoints. The results are strongest on objects whose geometry is difficult for compact global latents, including chairs with legs and armrests, drones with thin propeller supports, motorcycles with wheels and handles, and flowers with layered petals. These examples show that the proposed cross-axis slice representation provides enough local structure to reconstruct fine details while still producing globally coherent 3D assets.

\section{Ablations}\label{app:ablations}
\noindent\textbf{Topology Loss.}
Table~\ref{tab:abl_topo} studies the contribution of the slice-wise topology-preserving objective. Removing $\mathcal{L}_{\mathrm{topo}}$ weakens structural preservation because rendering losses alone do not explicitly penalize broken components, filled holes, or incorrect connectivity changes across slices. Using only $\mathcal{L}_{\mathrm{PH}}$ improves the topology of individual cross-sections, but does not directly constrain where topological events occur along the slicing direction. Conversely, using only $\mathcal{L}_{\mathrm{trans}}$ encourages event locations to align across depth, but provides weaker supervision for the detailed topology within each slice. The full objective combines these complementary signals and gives the best balance between surface fidelity and topological correctness. The ablation over $\lambda_{\mathrm{topo}}$ further shows that a moderate topology weight is preferable: too small a weight provides limited structural supervision, while too large a weight can over-constrain the decoder and reduce geometric fidelity. Increasing the number of supervised cross-sections improves the coverage of topological events, with the default $N_s{=}128$ providing the strongest supervision without changing the compact latent layout.
\begin{table}[t]
  \centering
  \small
  \setlength{\tabcolsep}{8pt}
  \renewcommand{\arraystretch}{1.05}
  \caption{\textbf{Ablation of the slice-wise topology-preserving loss.}
  All variants are evaluated on VAE reconstruction. Default settings are shaded.}
  \label{tab:abl_topo}
  \vspace{-0.2cm}
  \begin{tabular}{lccccc}
    \toprule
    \textbf{Variant}
    & \textbf{CD} $\downarrow$
    & \textbf{F@0.01} $\uparrow$
    & \textbf{F@0.005} $\uparrow$
    & \textbf{IoU} $\uparrow$
    & \textbf{Betti-Err} $\downarrow$ \\
    \midrule

    \rowcolor{SILSASection}
    \multicolumn{6}{l}{\textit{Loss components}} \\
    No $\mathcal{L}_{\mathrm{topo}}$
    & 0.78 & 93.84 & 78.16 & 81.76 & 4.43 \\
    Only $\mathcal{L}_{\mathrm{PH}}$
    & 0.72 & 94.51 & 79.72 & 84.41 & 2.91 \\
    Only $\mathcal{L}_{\mathrm{trans}}$
    & 0.69 & 95.03 & 80.46 & 87.16 & 2.87 \\
    \rowcolor{SILSARow}
    Full $\mathcal{L}_{\mathrm{topo}}$ (default)
    & 0.59 & 96.79 & 84.03 & 93.01 & 1.58 \\
    \midrule

    \rowcolor{SILSASection}
    \multicolumn{6}{l}{\textit{Topology loss weight $\lambda_{\mathrm{topo}}$}} \\
    $\lambda_{\mathrm{topo}}{=}0.01$
    & 0.66 & 95.62 & 82.14 & 90.37 & 2.24 \\
    \rowcolor{SILSARow}
    $\lambda_{\mathrm{topo}}{=}0.1$ (default)
    & 0.59 & 96.79 & 84.03 & 93.01 & 1.58 \\
    $\lambda_{\mathrm{topo}}{=}0.5$
    & 0.62 & 96.31 & 83.27 & 92.42 & 1.73 \\
    $\lambda_{\mathrm{topo}}{=}1.0$
    & 0.67 & 95.41 & 81.92 & 90.86 & 1.96 \\
    \midrule

    \rowcolor{SILSASection}
    \multicolumn{6}{l}{\textit{Number of cross-sections $N_s$ per axis}} \\
    $N_s{=}32$
    & 0.68 & 95.07 & 81.03 & 88.94 & 2.36 \\
    $N_s{=}64$
    & 0.63 & 96.02 & 82.75 & 91.48 & 1.87 \\
    \rowcolor{SILSARow}
    $N_s{=}128$ (default)
    & 0.59 & 96.79 & 84.03 & 93.01 & 1.58 \\
    \bottomrule
  \end{tabular}%
\end{table}
\begin{table}[t!]
  \centering
  \small
  \setlength{\tabcolsep}{8pt}
  \renewcommand{\arraystretch}{1.05}
  \caption{\textbf{Ablation on the number of canonical axes.} Default setting is shaded.}
  \label{tab:ablation_axes}
  \vspace{-0.2cm}
  \begin{tabular}{lcccccc}
    \toprule
    \textbf{Variant}
    & \textbf{Axes}
    & \textbf{$N$}
    & \textbf{\#Tokens}
    & \textbf{CD} $\downarrow$
    & \textbf{IoU} $\uparrow$
    & \textbf{Betti-Err} $\downarrow$ \\
    \midrule
    \rowcolor{SILSASection}
    \multicolumn{7}{l}{\textit{Single axis}} \\
    1 axis ($z$ only)            & $z$       & 384 & 384 & 1.34 & 76.43 & 5.92 \\
    1 axis ($z$ only)            & $z$       & 128 & 128 & 1.87 & 71.29 & 7.83 \\
    1 axis ($x$ only)            & $x$       & 384 & 384 & 1.41 & 75.18 & 6.18 \\
    1 axis ($x$ only)            & $x$       & 128 & 128 & 1.94 & 70.42 & 8.07 \\
    \midrule
    \rowcolor{SILSASection}
    \multicolumn{7}{l}{\textit{Two axes}} \\
    2 axes ($x,y$)               & $x,y$     & 192 & 384 & 0.81 & 88.46 & 2.74 \\
    2 axes ($x,z$)               & $x,z$     & 192 & 384 & 0.79 & 88.91 & 2.61 \\
    \midrule
    \rowcolor{SILSASection}
    \multicolumn{7}{l}{\textit{Three axes}} \\
    \rowcolor{SILSARow} 3 axes (default)             & $x,y,z$   & 128 & 384 & 0.59 & 93.01 & 1.58 \\
    \bottomrule
  \end{tabular}%
\end{table}

\noindent\paragraph{Number of Axes.}
Table~\ref{tab:ablation_axes} ablates the number of canonical slicing axes used by the SliceVAE while approximately matching the total token budget. Single-axis variants degrade substantially regardless of slicing direction: even when given $N{=}384$ slices to match the default token count, the model lacks cross-sectional evidence orthogonal to its slicing direction, leading to severe topological errors (Betti-Err of 5.92 for $z$-only and 6.18 for $x$-only). Reducing the budget to $N{=}128$ amplifies the gap further, suggesting that information lost during single-axis compression is not fully recovered by the decoder.
Two-axis variants close most of the gap, with $(x, z)$ slightly outperforming $(x, y)$ since orthogonal vertical and horizontal slicing captures more complementary structure for typical upright objects. However, both two-axis configurations still trail the full three-axis design, particularly on Betti-Err (2.61--2.74 vs.\ 1.58), suggesting that the third axis reinforces topological consistency by exposing structures that any two cross-sectional views jointly underdetermine.
 
\noindent\textbf{VAL Update Mechanism.}
Table~\ref{tab:ablation_val_update} ablates how slice tokens write to the Volumetric Anchor Lattice. Replacing gated writes with full overwrite causes each transformer block to clobber the accumulated cross-axis evidence with the latest token's contribution, breaking the multi-block coordination that makes VAL effective (CD 0.81, Betti-Err 2.74). Switching to additive writes preserves prior evidence but lets magnitudes accumulate without channel-wise selectivity, which improves over overwrite but still trails the gated variant (CD 0.75, Betti-Err 2.01). The default gated write achieves the best results across all metrics by allowing the model to learn, per channel, how much existing VAL content to retain versus replace as new slice evidence arrives. This selectivity is what enables the VAL to function as a stable shared workspace across transformer blocks rather than as a noisy buffer.
\begin{table}[t]
  \centering
  \small
  \setlength{\tabcolsep}{10.5pt}
  \renewcommand{\arraystretch}{1.05}
  \caption{\textbf{Ablation on the VAL update mechanism.} Image-conditioned generation from noise with the frozen SliceVAE decoder. Default setting is shaded.}
  \label{tab:ablation_val_update}
  \vspace{-0.2cm}
  \begin{tabular}{lcccc}
    \toprule
    \textbf{Variant}
    & \textbf{CD} $\downarrow$
    & \textbf{F@0.01} $\uparrow$
    & \textbf{IoU} $\uparrow$
    & \textbf{Betti-Err} $\downarrow$ \\
    \midrule
    Overwrite (no gating)
    & 0.81 & 94.01 & 86.32 & 2.74 \\
    Additive write (no gating)
    & 0.75 & 94.77 & 87.24 & 2.01 \\
    \rowcolor{SILSARow}
    Gated write (default)
    & 0.72 & 95.04 & 89.11 & 1.81 \\
    \bottomrule
  \end{tabular}%
  \vspace{-0.3cm}
\end{table}

\section{Discussion}
\noindent\textbf{3D Topology Preservation.}
Topology refers to the structural properties of a shape that remain invariant under continuous deformation: the number of connected components, the presence and count of holes, and the way these structures relate across the object. For 3D shapes, these properties are formalized through Betti numbers, where $\beta_0$ counts connected components, $\beta_1$ counts loops or tunnels, and $\beta_2$ counts enclosed voids. Unlike point-wise geometric metrics such as Chamfer distance or surface error, topological correctness captures whether a reconstructed shape preserves the qualitative structure of the original, whether a chair has four separable legs rather than three fused ones, whether a wheel retains its central opening rather than filling in, and whether a railing's bars remain distinct from one another while maintaining their intended connections to the frame. These distinctions matter because shapes with low surface error can still be structurally wrong: a generated mesh that fills a hole, breaks a thin support, or merges two nearby parts will register only small per-vertex deviations from the ground truth while fundamentally misrepresenting what the object is.

Persistent homology~\cite{edelsbrunner2002topological,zomorodian2004computing} provides a principled way to quantify and supervise these structural properties during learning. By tracking how topological features appear and disappear across a filtration of the shape — for example, sweeping a level set through an SDF — persistent homology produces a multi-scale signature that records each feature's birth, death, and persistence. Topological features that persist across a wide range of filtration values correspond to robust structures, while short-lived features correspond to noise. Loss functions based on persistent homology, Betti-number matching, or skeleton overlap have proven effective for 2D segmentation tasks involving thin or branching structures~\cite{hu2019topology,clough2020topological,shit2021cldice,stucki2023topologically}, and related work applies structural constraints to 3D implicit and mesh reconstruction~\cite{jignasu2024stitch,zhang2025topology}. However, applying full volumetric topology supervision to high-resolution 3D generation remains computationally prohibitive, since persistence diagrams must be recomputed across many decoded samples and at fine spatial resolution. SILSA addresses this by exploiting the slice-based structure of its latent representation: persistence is matched within individual cross-sections, and Betti transitions are aligned across neighboring slices, supervising topology at tractable per-slice cost while still capturing how connectivity evolves through the shape.

\noindent\textbf{Motivation for VAL.} Generating the three axis streams independently provides no direct mechanism for reconciling their geometry. Direct cross-attention enables communication between streams, but does not explicitly organize their interactions by shared 3D location. VAL uses the spatial arrangement of the slices: each token reads from and writes to its corresponding grid plane, allowing intersecting slices to exchange information through shared anchors. This introduces geometric structure into cross-axis communication and allows evidence to accumulate across transformer blocks.\looseness-1

\noindent\textbf{Editing Applications and Editability.} Because SILSA's slice latents are indexed by axis and depth, selected codes can be replaced or interpolated while leaving the remaining codes fixed. This provides a potential mechanism for shape composition and interactive geometry editing, such as transferring geometry between canonically aligned shapes within a selected depth range. Future extensions could also explore partial noising and image-conditioned denoising of selected latents. However, each latent summarizes an entire overlapping planar window, and the decoder fuses information across all three axes, so latent-level interventions do not guarantee spatially confined mesh changes. In particular, selecting slices from multiple axes does not by itself isolate their intersection, since each token contributes features across its entire grid plane. Developing reliable local and sub-slice editing procedures that maintain cross-axis consistency and preserve geometry outside the target region remains an interesting direction for future work.

\noindent\textbf{Limitations.}
Like most learning-based 3D generation methods, SILSA's performance depends on the diversity and scale of the training distribution, and objects with structural patterns far outside this distribution may be reconstructed with reduced fidelity. Generation quality is also influenced by the quality of the input image, with ambiguous or low-information views potentially yielding less faithful 3D structure. Our slices, topology supervision, and anchor assignments are defined in a fixed canonical coordinate frame, so the representation is not rotation-invariant. In addition, \modelnamenc{} currently generates geometry only, without predicting texture or appearance. Since the topology losses operate on canonical 2D cross-sections, they encourage but do not guarantee preservation of the final mesh's global 3D topology, particularly for oblique structures, tunnels, and enclosed cavities. Scaling to broader data sources is a promising direction for future work.

\noindent\textbf{Broader Impacts.}
Our method contributes to high-resolution image-to-3D generation, with positive applications in content creation, design, education, and simulation, where it lowers the barrier to producing 3D assets. The improved efficiency of our slice-based representation also makes high-resolution 3D generation more accessible to researchers with limited computational resources. As with other generative models, advances in 3D generation carry risks including unauthorized 3D replicas of real objects and the displacement of manual modeling tasks, which downstream applications should address through appropriate safeguards.



\end{document}